# AI for Green Spaces: Leveraging Autonomous Navigation and Computer Vision for Park Litter Removal

Christopher Y. Kao, Akhil Pathapati, James Davis

**Abstract**— There are 50 billion pieces of litter in the U.S. alone. Grass fields contribute to this problem because picnickers tend to leave trash on the field. We propose building a robot that can autonomously navigate, identify, and pick up trash in parks. To autonomously navigate the park, we used a Spanning Tree Coverage (STC) algorithm to generate a coverage path the robot could follow. To navigate this path, we successfully used Real-Time Kinematic (RTK) GPS, which provides a centimeter-level reading every second. For computer vision, we utilized the ResNet50 Convolutional Neural Network (CNN), which detects trash with 94.52% accuracy. For trash pickup, we tested multiple design concepts. We select a new pickup mechanism that specifically targets the trash we encounter on the field. Our solution achieved an overall success rate of 80%, demonstrating that autonomous trash pickup robots on grass fields are a viable solution.

## I. Introduction

Litter continues to be a significant problem around the world. Indeed, a 2020 study revealed that, in America alone, there are 50 billion pieces of litter [1]. Furthermore, this statistic has been increasing since 2010 [2].

There have been robotic solutions that tackle different aspects of litter, notably on beaches and water [3][4]. Beach-cleaning robots take advantage of sand's forgiving nature, allowing simple sifters to be implemented effectively. There are also water-cleaning robots which leverage water currents: the movement of water groups the trash together, removing the need for autonomous movement.

Grass terrain presents its own challenges. Grass is a living organism; sifting methods used in the sand would damage the grass. Indeed, Roomba-style continuous vacuuming and brushing would dig up the roots. Furthermore, litter is spread out in the field, and autonomous navigation is required to reach it.

There have been robotic solutions that work in conjunction with humans [5]. In this method, human workers still need to pick up the litter, thus making it an incomplete solution to litter clean up.

There have also been research that simulates trash collecting robots in a virtual environment. Such research tends to simplify the problem. One such article focuses only on the path planning aspect and does not propose a design for the robot [6]. Instead, it creates a 2D environment for the robot to traverse, assuming perfect sensor data. This does not account for sensor noise and variability of actions in the real world. Another article proposes a design for a trash collecting robot [7]. However, the robot is only simulated in Solidworks. It was neither constructed, nor field tested.

Autonomous cleaning robots on land have also been proposed. For instance, one cleaning robot leverages ultrasonic sensors to detect trash [8]. However, the design would not function in grass fields. Furthermore, the ultrasonic sensor would easily mistake obstacles for trash,

Previous research on using autonomous robots on grass fields to pick up litter has been limited. A prior article proposed a robot which utilizes a 5 degrees-of-freedom arm and a ResNet Convolutional Neural Network (CNN) that does image segmentation [9]. The robot differentiates between different types of trash, meaning it can only detect the types of trash it has been trained on. Furthermore, the path planning of the robot is limited, only allowing the robot to cover rectangular areas.

This paper expands on current research, proposing a robot that can pick up a wide variety of trash, and cover a grass field of any shape and size with maximum efficiency. To accomplish this, we draw upon research done in autonomous navigation, path planning, and image classification.

The most versatile algorithm for path planning is simultaneous localization and mapping (SLAM). This method is utilized when the robot must actively gather information about and traverse the area at the same time [10]. This algorithm is prevalent in room cleaning robots but does not produce the optimal path: the robot will double back on itself whilst exploring the area. Instead, we chose a different solution called Spanning Tree Covering (STC) algorithm. When given some information about the area's shape and size, the algorithm first divides the area to cover into cells corresponding to the detection area. Then, it follows a tree that spans the entire graph, covering the area only once [11].

There has also been significant research done in using image segmentation and CNNs to detect trash [12][13]. The Trash Annotations in Context (TACO) dataset is used for training such models due to its varied environments and types of trash [14]. This approach, however, does not meet the needs of our project. By using image segmentation, the models are limited to the 60 classes of trash in the TACO dataset. Furthermore, these research use artificial backgrounds to highlight the trash when testing the model [12]. In contrast, we aim to detect a wider variety of trash beyond the TACO dataset and detect on a natural background. To accomplish this, we decided on an image classification model, with the

Christopher Kao and Akhil Pathapati are with 77SPARX Studio, Inc., Sunnyvale, CA 94087 USA (e-mail: ckao@77sparx.com, apathapati@77sparx.com).
Professor James Davis is with the Department of Computer Science, University of California Santa Cruz, Santa Cruz, CA 95064 USA (e-mail: davis@cs.ucsc.edu).

two classes being "trash" and "no-trash". We also collected our own dataset of 18,154 images to ensure that the model performs well.

Lastly, there has been much research done in robotic pickup methods. The research is focused on precise pickup mechanisms that can pick up single items at a time. These methods typically use a robotic arm to manipulate an end effector [9]. As for the end effector, research has been done in soft grippers, which can grab a wider range of items. One such end effector is the granular jammer, which deforms around the item to pick it up [15]. However, this method relies on a solid edge or corner to deform around; this makes the mechanism ineffective against flimsy pieces of plastic litter. We tested multiple designs before choosing a pickup mechanism inspired by the bow rake. By brushing an area of the grass, it can pick up trash that rests on the grass. This takes advantage of the lightweight nature of the trash we are picking up to perform better than other methods.

## II. METHODOLOGY

### A. Robot Structure

The main structure of our robot must be adaptable as components and modules are added to the robot while still providing a stable structure on which to build the robot from. We decided to pursue a prefabricated robot chassis from SuperDroid Robots to house the drive motors, which are REV Neo 775 Pro motors controlled with REV Spark MAX Motor Controllers. We utilized T-Slots to build shelves and housing for the other components and mechanisms.

The main robot processor is the roboRIO. We also utilized the Arduino ecosystem as an interface between the roboRIO and components including limit switches, linear actuators, GPS module, and a computer co-processor.

### B. Autonomous Navigation

There are several criteria the autonomous navigation solution must meet. First, the solution must limit the robot to only the coverage area. Many grass fields have fences surrounding them; any attempted movement out of the field would result in a crash. Furthermore, the solution should avoid any obstacles in the coverage area, such as trees, rocks, and signs. Finally, the field traversal should not double back on itself. This ensures that we minimize the amount of time it takes to traverse the field. Accomplishing these tasks will require the robot to know its location and orientation in the field. The following paragraphs will present our approach.

*1) Path Planning*

Optimized path planning plays a crucial role in having an efficient robot. The algorithm we decided to use is the STC algorithm [11]. It is a general yet methodical solution to path planning when given only the vertices of the area and obstacles. As shown in Figure 1, the robot implements the algorithm as follows: 1) The robot is given the vertices of the area it needs to cover, and vertices of any obstacles in the area. 2) The robot splits the area to cover into a grid of 2 x 2-unit squares, with a unit square (0.3 m x 0.3 m) representing the size of the detection zone. 3) Generate a tree using the grid of 2 x 2-unit squares. This tree represents the "walls" from which the robot will plan its path. 4) From the starting point, which is where the robot is positioned initially, the path is created by keeping the wall on the path's right side until it gets back to the starting position. Once this is done, the entire area will have been covered once.

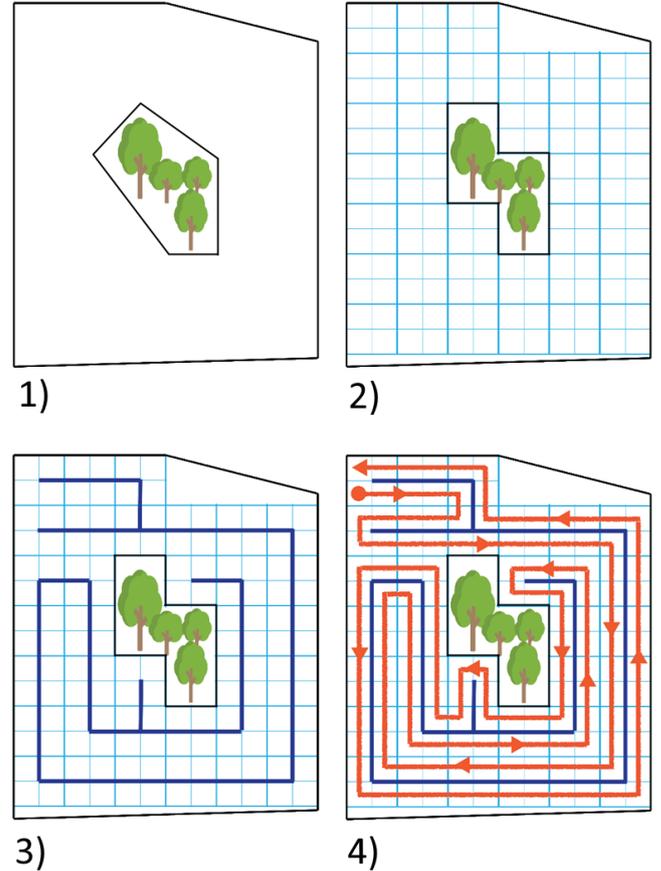

Fig. 1: Visualization of STC Algorithm. In 2), the thick blue lines mark the 2 x 2-unit squares, while the thin blue lines mark the unit squares. In 3), the dark blue lines mark the tree generated from the algorithm. In 4), the red lines mark the path the robot follows.

*2) Localization*

To perform autonomous navigation, we need precise and accurate positioning data. The common solution is Global Positioning System (GPS). However, GPS itself is only accurate within a 4.9-meter radius in optimal conditions [16]. This is too coarse to navigate safely and accurately; if the robot is adjacent to a fence, it risks running through the barrier. However, combining Real-Time Kinematic (RTK) with GPS allows the robot to achieve location accuracy down to 0.01 m. As shown in Figure 2, RTK-GPS uses a separate base station whose exact location is known. When the base station receives a location estimate from the GPS signal, it calculates the error in that estimate. If the base station is close enough to the robot or rover—less than 10,000 m away—the location error of the base station can be used to correct the rover's location error down to 0.01 m [17][18][19]. In addition, there are many different RTK networks with widespread coverage, making RTK-GPS an easy and inexpensive solution [20].

To implement RTK-GPS, we used a ZED-F9P breakout board that is part of the Arduino Ecosystem. It gives the robot an RTK-GPS signal every second [21].

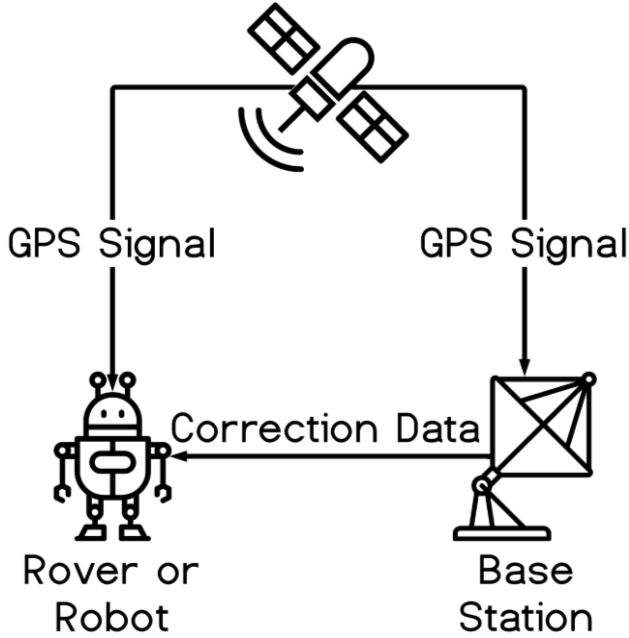

Fig. 2: RTK-GPS Implementation Diagram. The corrections calculated by the base station (reference station) are sent to the robot, which uses it to correct the GPS data of its own location.

Due to the slow update frequency of the GPS, we also used a gyroscope and dead reckoning to estimate our location between GPS signals. We used the navX2-MXP Inertial Navigation Unit (IMU) board as our gyroscope. During testing, we observed yaw drift which caused inaccuracies in the robot's position estimation. To rectify this, our system uses two GPS signals that are recorded apart from each other when the robot is driving in a straight line. From these readings, our system uses the following equation to calculate and correct the error from the IMU:

$$\Delta Y = \arctan\left(\frac{\Delta \phi}{\Delta \lambda}\right) - Y$$

Where $\Delta Y$ is the error, $\Delta \phi$ is the change in longitude, $\Delta \lambda$ is the change in latitude, and $Y$ is the estimate from the IMU.

To test this algorithm, Figure 3 shows the graph of two different autonomous navigation trials: A) is the path when the robot is not using our correction algorithm, and B) is the path when the robot is. As shown, our yaw drift correction algorithm significantly boosts the accuracy of the robot. In A), the robot's location estimate (orange) does not line up with the calculated path (blue). Furthermore, there is a jump every so often of the orange line, signaling that the robot has a poor estimate of its location, which is drastically corrected every time the robot receives a GPS signal of its true location. On the other hand, B) depicts a trial using the yaw drift correction algorithm. In contrast to A), in B), the robot's estimated location lines up closely with the target path, meaning the robot is following the intended path much more closely. In addition, there are far fewer jumps in the orange path, meaning the robot's location estimate is much closer to its true location.

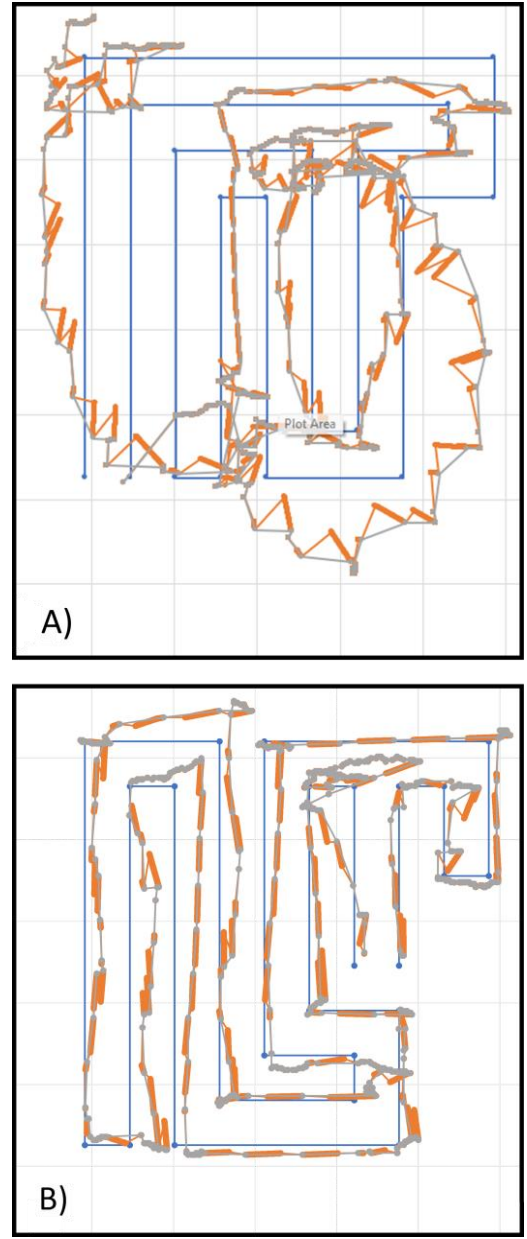

Fig. 3: Graph of robot path without calibration (A) vs. with calibration (B). The total area is 7.701 square meters. In both graphs, the blue line is the path to follow; the gray nodes are the locations where the robot receives a GPS signal; and the orange nodes are the estimated locations between GPS signals.

### C. Computer Vision Machine Learning Trash Detection

Trash identification is an integral part of the project. To be able to detect any type of trash, we utilize research in CNNs. To ensure we could detect all types of trash, we use image classification instead of segmentation, defining two classes—trash and no-trash. By classifying into these two categories, the model learns to generalize the characteristics of trash to a variety of shapes and sizes.

We first need a feasible way to gather thousands of pictures of trash and grass. There are existing trash datasets, notably TACO Dataset [14]. However, this dataset is small, with only 1500 annotated images. Furthermore, TACO contains a wide variety of backgrounds, whereas our problem focuses on a grass background. Consequently, we decided to create our own dataset. To do so, we created a python script that utilized the camera mounted on our robot to rapidly take frames from the video feed. By predefining what the camera is looking at—trash or no-trash—the robot puts the images into their respective folder, without human labeling. Next, we augmented the data using the techniques outlined in Table I. This was to account for the varied grass color and texture of different local fields: including sparse blades, dense clumps, and yellow dried out fields as shown in Figure 4. Through these techniques, we reached 18,154 images, which were split into a training and validation dataset. The test images were sampled from a different dataset, which we gathered from a different location than where the training and validation dataset was collected. The testing dataset has 6,093 images, with 2,920 'no-trash' images and 3,173 'trash' images.

TABLE I: Data augmentation techniques.

| Augmentation Used | Description |
|---|---|
| Rotation | To generate more than one training data point from one image |
| Hue | To account for different colors of grass, including lush grass and dried yellow grass |
| Brightness | To account for testing in different lighting conditions |
| Saturation | To account for quality of lighting & weather conditions. |

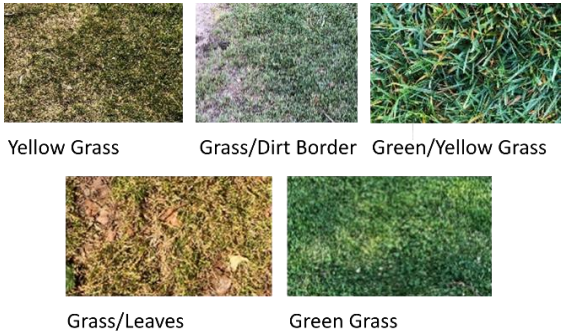

Fig. 4: Variable grass conditions seen in parks.

Next, we trained three different models: EfficientNet, MobileNet, and ResNet50. We compared the validation accuracy and test accuracy of each model, as shown in Table II. While all three models had >96% validation accuracy, the ResNet50 model performed better on the test dataset with a 94.52% test accuracy, compared to the 83.99% and 92.71% accuracy of the MobileNet and EfficientNet respectively. From these results, we decided to implement the ResNet50 model for our project.

TABLE II: Accuracies of different models

| Model Name | Validation Accuracy | Test Accuracy |
|---|---|---|
| MobileNet | 96.53% | 83.99% |
| EfficientNet | 96.61% | 92.71% |
| ResNet50 | 96.36% | 94.52% |

The flowchart of the trash detection algorithm is shown in Figure 5. A webcam gives a live feed of what is in front of the robot. Each frame is fed to the python trash detection script running on the laptop connected to the webcam. The frame is then downsized to 300 x 300 pixels to increase inference speed; the ResNet50 model makes an inference from this frame. Since false positives are better than false negatives (rather excessively pickup than miss pieces of trash), we set the threshold for inferring 'trash' to 50%. The prediction is then sent to the roboRIO. To minimize false positives caused by this low threshold, the roboRIO takes the moving average of 10 predictions. If the average is more than 90% trash, the robot begins the trash pickup mechanism.

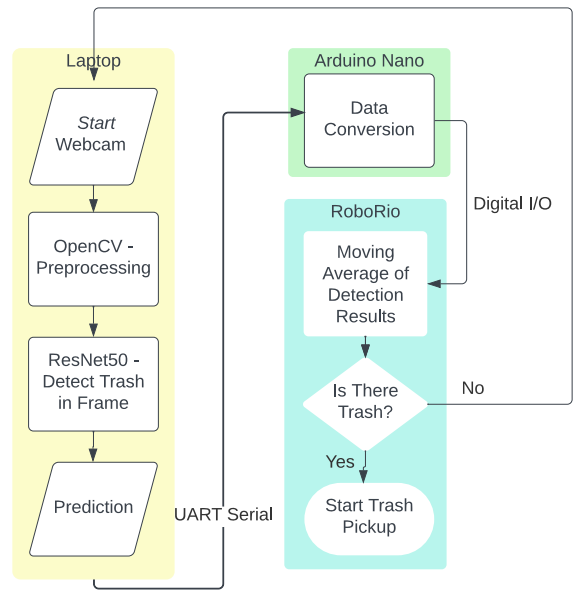

Fig. 5: Computer vision trash identification flowchart.

### D. Trash Pickup Mechanism

Autonomously picking up litter is a hard task, with many prior papers using elaborate techniques such as origami design [22], Fin-Ray effect gripper [23], and granular jammer [15]. To select the best pickup design for our task, we tested existing designs as well as common household methods. For each mechanism, we simulated a robotic movement. We used 10 different pieces of trash from local fast-food places against each pickup method. For each piece of trash, we had 4 pickup attempts. The result of our testing is shown in Table III.

From our results, we found the most promising pickup method to be the rake pickup. It picks up trash in an area rather than singular pieces of trash, which would be more effective to implement. The rake picks up trash when the tines are parallel to the ground. This allows part of the tines to skirt the grass like a comb through hair. This leaves the grass alone but will pick up trash resting on top of the grass. Once the trash is picked up, however, the mechanism cannot move the litter to a compartment inside the robot. To solve this, we used the brush in junction with the rake pickup concept to sweep the trash picked up into a receptacle in the robot.

TABLE III: Trash pickup methods tested.

| Image | Accuracy | Comments |
|---|---|---|
| 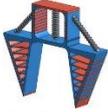 Fin-Ray Grabber | 65.0% | The Fin-Ray effect gripper [23] is inspired by the physiology of fish fins, gripping more around the object. While the Fin-Ray effect gripper does work well for solid objects, there was no advantage to other established pickup methods, like the Reacher Grabber. |
| 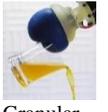 Granular Jammer | 12.5% | The granular jammer [15] has three parts: the granules, boundary layer membrane, and a vacuum pump. Once the membrane filled with granules is pressed against an object, the vacuum removes air and locks the granules in place. This creates the grip force. The granular jammer relies on solid protrusions on the object to grip on and hold. Without this, it won't work. Thus, the granular jammer works poorly on soft trash, like bags, wrappers, and flat boxes. |
| 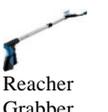 Reacher Grabber | 70.0% | The Reacher Grabber [24] is the go-to gripper for most. It is quite simple, with a single tendon closing a two-point claw when tensioned. It performs comparably to the Fin-Ray Gripper. The main concern is that it can only pick up from one place at a time. This means the mechanism would not be feasible without image segmentation. |
| 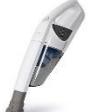 Vacuum | 40.0% | The vacuum has a 0.02 m x 0.04 m opening which sucks up anything directly below it. This makes it effective with small objects. However, bigger, soft trash will clog the opening, making the vacuum inoperable. |
| 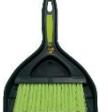 Brush & Dustpan | 65.0% | The brush and dustpan work by sliding the brush towards the dustpan, usually in an elliptical motion. For the sake of simplicity, we used a lateral motion parallel to the ground. It works quite well, picking up all sorts of trash. The only drawback is it tends to uproot the grass. |
| 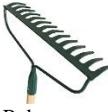 Rake (Tines) | 82.5% | The bow rake works by sliding in the grass, like a comb through hair. Since the trash is light, it would rest on top of the grass. We would slide underneath and pick it up. This method also works quite well for light pieces of litter, which we encounter often. |

## III. RESULTS

We tested our completed robot, shown in Figure 6, against 10 commonly found pieces of trash at the park, shown in Figure 7. These pieces of trash are of different varieties to ensure a thorough and accurate representation of the research. The varied geometry of the trash demonstrates our robot's capability to pick up any piece of trash, no matter shape or size. Furthermore, to demonstrate the ability of our algorithm to detect any type of trash, half of the trash we used in the final test are not in the dataset used to train the model.

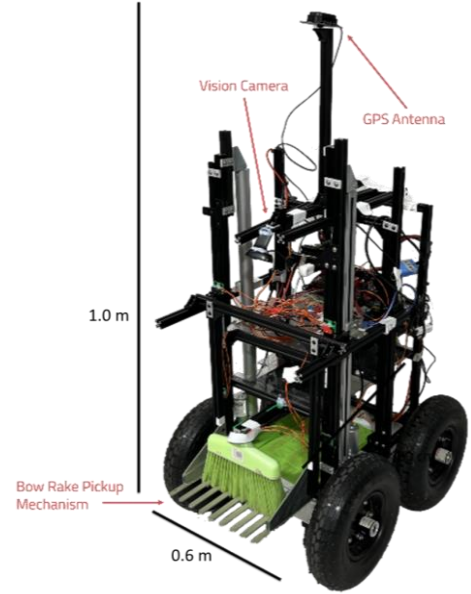

Fig. 6: Final robot design. The camera sees a 0.3 m x 0.3 m square in front of the robot. If there is trash in that area, then the green pickup mechanism is deployed. The GPS Antenna is placed at the very top to ensure a strong signal.

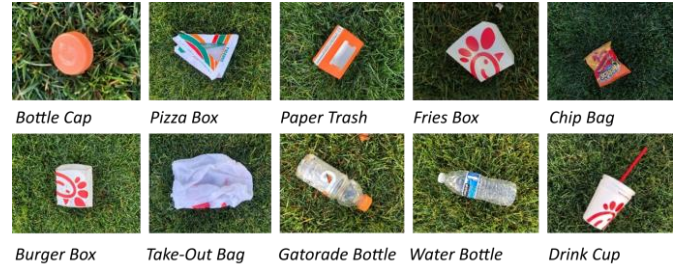

Fig. 7: Trash used in final test

To evaluate the autonomous navigation solution, we tested it in our local park. The coverage area is a quadrilateral shape with an area of 44.87 square meters, and the robot cleaned the area in 16 minutes and 47 seconds. We also included an obstacle defined by 3 vertices in the center of the field to demonstrate the robot's ability to avoid static obstacles. We evaluated the success of each aspect of the robot in the following manner:

- **Autonomous Navigation** is rated by the proportion of trash the robot drives over (e.g. If 85% of trash is driven over by the robot, then the Autonomous Navigation success Rate would be 85%).
- **Trash Detection** success rate is the proportion of trash detected by the robot given that the trash was driven over by the robot.
- **Trash Pickup** success rate is the proportion of trash collected given the trash was detected. The success rates from our test are shown in Tabel IV.

TABLE IV: Success rate of different aspects of the robot.

| Aspect of Robot | Success Rate |
|---|---|
| Autonomous Navigation | 100% |
| Trash Detection | 90% |
| Trash Pickup | 89% |
| Total | 80% |

## IV. Discussion

Overall, the robot covered in this research paper performed well for the target task.

The autonomous navigation method we used worked without error. The STC algorithm we leveraged was simple and efficient, properly covering the field, no matter what size or shape. The RTK-GPS gave the robot centimeter-level accuracy, allowing the robot to properly follow the generated path. Together, this method enabled the robot to navigate to every single piece of litter.

The computer vision trash detection model had satisfactory performance. It was able to detect pieces of trash that it did not directly train on and maintain a high accuracy. Because we set the threshold of a 'trash' prediction low, the model predicted false positives which were caused by change in lighting from shadow. This did not impede the accuracy of the robot, though it did slow it down by initiating unneeded pickup cycles. This could be solved by increasing the confidence threshold at which to predict 'trash' or collect images specifically of shadows to train the model to ignore such variations. The model also failed to detect the small bottle cap. However, bottle caps by themselves are not common—most are still fastened to the plastic bottle—so this is an unlikely case to run into [1].

Lastly, the trash pickup mechanism performed at par for the target pieces of trash. The light plastic and paper trash was exactly what our pickup method was designed for; accordingly, it picked up the plastic bottles and wrappers with ease. The trash it struggled with was the paper bag, which was slightly too big for our pickup method to collect. This can be remedied by enlarging the pickup mechanism. This pickup method stands to be a new and effective solution.

## V. Conclusion

We designed a robot for picking up trash in grass fields. Our trash mostly consists of plastic water bottles, wrappers, and boxes that are left from picnickers. The robot must be able to autonomously navigate, detect, and pick up trash. As shown by our research, such a robot is feasible, and picks up 80% of trash. We believe this will relieve the strain on current park maintenance workers and make our parks nicer to visit for everyone.

## Acknowledgment

The authors would like to express their gratitude to Aayush Gaywala for his support and assistance during this project.

## References

[1] "Litter-Study-Summary-Report-May-2021_final_05172021.pdf." Accessed: Jul. 22, 2024. [Online]. Available: https://kab.org/wp-content/uploads/2021/05/Litter-Study-Summary-Report-May-2021_final_05172021.pdf

[2] O. US EPA, "Report on the Environment (ROE)," US EPA. Accessed: Jul. 31, 2024. [Online]. Available: https://www.epa.gov/report-environment

[3] "BeBot | The beach cleaning robot," The Searial Cleaners. Accessed: Jul. 22, 2024. [Online]. Available: https://searial-cleaners.com/our-cleaners/bebot-the-beach-cleaner/

[4] "Collec'Thor | Fixed trash collector," The Searial Cleaners. Accessed: Jul. 22, 2024. [Online]. Available: https://searial-cleaners.com/our-cleaners/collecthor-the-fixed-waste-collector/

[5] K. Bredesen, H. Arnarson, B. Solvang, and A. Anfinnsen, "Human-robot collaboration for automatic garbage removal," in *2022 IEEE/SICE International Symposium on System Integration (SII)*, Jan. 2022, pp. 803–808. doi: 10.1109/SII52469.2022.9708825.

[6] C.-H. Chiang, "Vision-based coverage navigation for robot trash collection task," in *2015 International Conference on Advanced Robotics and Intelligent Systems (ARIS)*, May 2015, pp. 1–6. doi: 10.1109/ARIS.2015.7158229.

[7] M. Kulshreshtha, S. S. Chandra, P. Randhawa, G. Tsaramirsis, A. Khadidos, and A. O. Khadidos, "OATCR: Outdoor Autonomous Trash-Collecting Robot Design Using YOLOv4-Tiny," *Electronics*, vol. 10, no. 18, Art. no. 18, Jan. 2021, doi: 10.3390/electronics10182292.

[8] "Garbage Collector Robot," Florida Atlantic University. Accessed: Aug. 09, 2024. [Online]. Available: https://www.fau.edu/engineering/senior-design/projects/fall2020/garbage-collector-robot/index.php

[9] J. Bai, S. Lian, Z. Liu, K. Wang, and D. Liu, "Deep Learning Based Robot for Automatically Picking Up Garbage on the Grass," *IEEE Trans. Consum. Electron.*, vol. 64, no. 3, pp. 382–389, Aug. 2018, doi: 10.1109/TCE.2018.2859629.

[10] B. Liu, Z. Guan, B. Li, G. Wen, and Y. Zhao, "Research on SLAM Algorithm and Navigation of Mobile Robot Based on ROS," in *2021 IEEE International Conference on Mechatronics and Automation (ICMA)*, Aug. 2021, pp. 119–124. doi: 10.1109/ICMA52036.2021.9512584.

[11] Y. Gabriely and E. Rimon, "Spanning-tree based coverage of continuous areas by a mobile robot," in *Proceedings 2001 ICRA. IEEE International Conference on Robotics and Automation (Cat. No.01CH37164)*, May 2001, pp. 1927–1933 vol.2. doi: 10.1109/ROBOT.2001.932890.

[12] E. Almanzor, N. R. Anvo, T. G. Thuruthel, and F. Iida, "Autonomous detection and sorting of litter using deep learning and soft robotic grippers," *Front. Robot. AI*, vol. 9, Dec. 2022, doi: 10.3389/frobt.2022.1064853.

[13] T. Wang, Y. Cai, L. Liang, and D. Ye, "A Multi-Level Approach to Waste Object Segmentation," *Sensors*, vol. 20, no. 14, Art. no. 14, Jan. 2020, doi: 10.3390/s20143816.

[14] P. F. Proença and P. Simões, "TACO: Trash Annotations in Context for Litter Detection," Mar. 17, 2020, *arXiv*: arXiv:2003.06975. doi: 10.48550/arXiv.2003.06975.

[15] A. C. Jacob and E. L. Secco, "Design of a Granular Jamming Universal Gripper," in *Intelligent Systems and Applications*, K. Arai, Ed., Cham: Springer International Publishing, 2022, pp. 268–284. doi: 10.1007/978-3-030-82199-9_16.

[16] "GPS.gov: GPS Accuracy." Accessed: Jul. 29, 2024. [Online]. Available: https://www.gps.gov/systems/gps/performance/accuracy/#how-accurate

[17] "RTK and the Federal Communications Commission (FCC) | GEOG 862: GPS and GNSS for Geospatial Professionals." Accessed: Jul. 29, 2024. [Online]. Available: https://www.e-education.psu.edu/geog862/node/1845

[18] "Placing the base | Reach RS/RS+." Accessed: Jul. 29, 2024. [Online]. Available: https://docs.emlid.com/reachrs/ppk-quickstart/placing-the-base/

[19] "Real Time Kinematic System For Precision Farming." Accessed: Jul. 29, 2024. [Online]. Available: https://archive.ph/20120203041216/http:/www.agnav.com/RealTimeKinematicSystem

[20] "Geodnet RTK Network." Accessed: Jul. 31, 2024. [Online]. Available: https://rtk.geodnet.com/

[21] "SparkFun GPS-RTK2 Board - ZED-F9P (Qwiic) - GPS-15136 - SparkFun Electronics." Accessed: Jul. 29, 2024. [Online]. Available: https://www.sparkfun.com/products/15136

[22] S. Li *et al.*, "A Vacuum-driven Origami 'Magic-ball' Soft Gripper," *Prof Rus*, May 2019, Accessed: Jul. 29, 2024. [Online]. Available: https://dspace.mit.edu/handle/1721.1/120930

[23] W. Crooks, G. Vukasin, M. O'Sullivan, W. Messner, and C. Rogers, "Fin Ray® Effect Inspired Soft Robotic Gripper: From the RoboSoft Grand Challenge toward Optimization," *Front. Robot. AI*, vol. 3, Nov. 2016, doi: 10.3389/frobt.2016.00070.

[24] "Amazon.com: Reacher Grabber Tool, 31" Grabbers for Elderly, Lightweight Extra Long Handy Trash Claw Grabber, Mobility Aid Reaching Assist Tool for Trash Pick Up, Nabber, Litter Picker, Arm Extension : Health & Household." Accessed: Aug. 15, 2024. [Online]. Available: https://www.amazon.com/Reacher-Foldable-Lightweight-Reaching-Extension/dp/B078RMCFWQ